# Piecewise Training for Undirected Models

**Charles Sutton and Andrew McCallum**
Computer Science Dept.
University of Massachusetts
Amherst, MA 01003
{casutton,mccallum}@cs.umass.edu

## Abstract

For many large undirected models that arise in real-world applications, exact maximum-likelihood training is intractable, because it requires computing marginal distributions of the model. Conditional training is even more difficult, because the partition function depends not only on the parameters, but also on the observed input, requiring repeated inference over each training example. An appealing idea for such models is to independently train a local undirected classifier over each clique, afterwards combining the learned weights into a single global model. In this paper, we show that this *piecewise* method can be justified as minimizing a new family of upper bounds on the log partition function. On three natural-language data sets, piecewise training is more accurate than pseudolikelihood, and often performs comparably to global training using belief propagation.

## 1 INTRODUCTION

Large graphical models are becoming increasingly common in applications including computer vision, relational learning [14], and natural language processing [17, 3]. Often the cheapest way to build such models is to estimate their parameters from labeled training data. But exact maximum-likelihood estimation requires repeatedly computing marginals of the model distribution, which is intractable in general.

This problem is especially severe for conditional training. If our final task is predict certain variables $\mathbf{y}$ given observed data $\mathbf{x}$, then it is appropriate to optimize the conditional likelihood $p(\mathbf{y}|\mathbf{x})$ instead of the generative likelihood $p(\mathbf{y},\mathbf{x})$. This allows inclusion of rich, overlapping features of $\mathbf{x}$ without needing to model their distribution, which can greatly improve performance [8]. Conditional training can be expensive, however, because the partition function $Z(\mathbf{x})$ depends not only on the model parameters but also on the input data. This means that parameter estimation requires computing (or approximating) $Z(\mathbf{x})$ for each training instance for each iteration of a numerical optimization algorithm; this can be expensive even if the graph is a tree.

To train such large models efficiently, an appealing idea is to divide the full model into pieces which are trained independently, combining the learned weights from each piece at test time. We call this *piecewise training*.

In this paper, we present a systematic evaluation of this intuitively-appealing procedure. We justify piecewise training as minimizes an upper bound on the exact log partition function. Although this bound can be proved directly, it can also be derived from the variational upper bounds presented by Wainwright, Jaakkola, and Willsky [15], a connection which motivates several generalizations of the basic piecewise procedure.

The piecewise estimator is also closely related to pseudolikelihood [1, 2]. Both estimators are based on locally normalizing small pieces of the full model. But pseudolikelihood conditions on the true value of neighboring nodes, which has the effect of coupling parameters in neighboring pieces (see Figure 3), while the piecewise estimator optimizes each piece independently. So the piecewise estimator is distinct from pseudolikelihood. Even though the difference may seem small, we show experimentally that the piecewise estimator is more accurate.

On three real-world natural language tasks, we show that the accuracy of piecewise training is often comparable to exact training. We also show that piecewise training performs better than pseudolikelihood, even if the pseudolikelihood objective is augmented to normalize over edges rather than single nodes.

These results suggest that piecewise training is preferable to pseudolikelihood as a method of choice for local training, allowing efficient training of massive real-world models where conditional training is currently impossible.

## 2 MARKOV RANDOM FIELDS

In this section, we briefly give background and notation on Markov random fields (MRFs) and conditional random fields (CRFs). A *Markov random field* is a probability distribution over a vector $\mathbf{y}$ that has been specified in terms of local factors $\psi$ as:

$$p(\mathbf{y}) = \frac{1}{Z} \prod_{st} \psi(y_s, y_t), \quad (1)$$

where the partition function $Z = \sum_{\mathbf{y}'} \prod_{st} \psi(y'_s, y'_t)$ normalizes the distribution. The distribution $p(\mathbf{y})$ can also be described as an undirected graphical model $\mathcal{G}$ with edge set $E = \{(s,t)\}$.

We assume that each of the local functions $\psi$ can be written in terms of weights $\boldsymbol{\theta}$ and functions $\phi$ as

$$\psi(y_s, y_t) = \exp\left\{\sum_\alpha \theta_{st;\alpha} \phi_{st;\alpha}(y_s, y_t)\right\}. \quad (2)$$

The functions $\phi_{st;\alpha}$ are the sufficient statistics of the model. For example, if the sufficient statistics are indicator functions of the form

$$\phi_{st;\alpha}(y_s, y_t) = 1_{\{y_s = y'_s\}} 1_{\{y_t = y'_t\}}, \quad (3)$$

then $\psi(y_s, y_t)$ is a lookup table where each value is $\psi(y_s, y_t) = \exp\{\theta_{st;y_s,y_t}\}$.

This choice of parameterization for the local factors ensures that the set $\{p(\mathbf{y}; \boldsymbol{\theta})\}$ is an exponential family. Letting $\alpha$ index over the exponential parameters over all edges $st$:

$$p(\mathbf{y}) = \exp\left\{\sum_\alpha \theta_\alpha \phi_\alpha(y_s, y_t) - A(\boldsymbol{\theta})\right\}, \quad (4)$$

where $A(\boldsymbol{\theta}) = \log Z$ normalizes the distribution.

Parameter estimation for MRFs can be done by maximum likelihood, but this requires computing $A(\boldsymbol{\theta})$, which is intractable. It is for this reason that approximations and bounds of $A$ are of great interest.

To simplify the exposition, we have assumed that the local functions are over pairs of variables. All of the discussion in this paper can easily be generalized to factors of higher arity.

A *conditional random field* is a Markov random field used to model the conditional distribution $p(\mathbf{y}|\mathbf{x})$ of target variables $\mathbf{y}$ given input variables $\mathbf{x}$. As above, let $\mathcal{G}$ be an undirected graph over $\mathbf{y}$ with edges $E = \{(s,t)\}$. Then a CRF models the conditional distribution as

$$p(\mathbf{y}|\mathbf{x}) = \exp\left\{\sum_{st}\sum_k \lambda_k f_k(y_s, y_t, \mathbf{x}) - A(\Lambda; \mathbf{x}).\right\}, \quad (5)$$

where $f_k$ are *feature functions* that can depend both on an edge in $\mathbf{y}$ and (potentially) the entire input $\mathbf{x}$, and $\Lambda = \{\lambda_k\}$ are the real-valued model parameters. Because the distribution over $\mathbf{x}$ is not modeled, the feature functions $f_k$ are free to include rich, overlapping features of the input without sacrificing tractability. Indeed, this is the chief benefit of using a conditional model.

For any fixed input $\mathbf{x}$, the distribution $p(\mathbf{y}|\mathbf{x})$ is an MRF with parameters

$$\theta_{st;y_s,y_t} = \sum_k \lambda_k f_k(y_s, y_t, \mathbf{x}), \quad (6)$$

and the indicator functions as sufficient statistics. We call this MRF the *unrolled graph* of the CRF for the input $\mathbf{x}$.

Parameter estimation in CRFs is performed by maximizing the log likelihood of fully-observed training data $\mathcal{D} = \{(\mathbf{x}^{(i)}, \mathbf{y}^{(i)})\}$, which is given by

$$\ell(\Lambda) = \sum_i \sum_{st} \sum_k \lambda_k f_k(y_s^{(i)}, y_t^{(i)}, \mathbf{x}^{(i)}) - \sum_i A\left(\mathbf{x}^{(i)}; \Lambda\right).$$

This is a convex function that can be maximized numerically by standard techniques, including preconditioned conjugate gradient and limited-memory BFGS. Quadratic regularization (i.e., a Gaussian prior on parameters) is often used to reduce overfitting.

Although *inference* for CRFs is thus exactly as in MRFs, *training* is more expensive. This is because the CRF log partition function $A(\Lambda; \mathbf{x})$ depends not only on the parameters but also on the input. Thus maximum-likelihood parameter estimation involves computing or approximating $A(\Lambda; \mathbf{x})$ once for each training instance for each iteration of a gradient ascent procedure. This can be expensive even when the unrolled graph is a tree.

## 3 PIECEWISE TRAINING

In this section, we present the piecewise estimator, justifying it as minimizing an upper bound on the log partition function. First, we present an example of a particularly intuitive case of piecewise estimation. Suppose we want to train a loopy pairwise MRF. In piecewise estimation, we simply train the parameters of each edge independently, as if each edge were a separate two-node MRF of its own. We take the learned parameters from this local training as the piecewise-trained edge parameters in the global model.

Now we define the piecewise estimator in more generality. We assume that the sufficient statistics of the distribution are partitioned into a set $\mathcal{P}$ of disjoint pieces; each piece $R \in \mathcal{P}$ is a set of integers indexing the sufficient statistics contained in piece $R$. For example, in a discrete pairwise MRF with tabular factors, we might choose each piece to correspond to the parameters and sufficient statistics for one edge factor in the MRF.

Then we define the piecewise objective function as

$$\ell_{\text{PW}}(\boldsymbol{\theta}) = \sum_{R \in \mathcal{P}} \sum_{\alpha \in R} \theta_\alpha \phi_\alpha(\mathbf{x}_\alpha) - \sum_{R \in \mathcal{P}} A_R(\boldsymbol{\theta}), \quad (7)$$

where $A_R(\boldsymbol{\theta})$ is the *local log partition function* for the piece, that is, $A_R(\boldsymbol{\theta}) = \log \sum_{\mathbf{x}_R} \exp\{\sum_{\alpha \in R} \theta_\alpha \phi_\alpha(\mathbf{x}_\alpha)\}$, where $\mathbf{x}_R$ is the vector of variables used anywhere in piece $R$. Finally, the piecewise estimator is defined as $\hat{\boldsymbol{\theta}}_{\text{PW}} = \max_{\boldsymbol{\theta}} \ell_{\text{PW}}$.

Consider the special case of per-edge pieces in a pairwise MRF. Then, for an edge $(s,t)$, we have $A_{st}(\theta) = \log \sum_{x_s, x_t} \psi(x_s, x_t)$, so that the piecewise estimator corresponds exactly to training independent probabilistic classifiers on each edge.

Now we make a few general remarks about this estimator. First, observe that the first summation in Equation 7 contains exactly the same terms the first summation in the exact likelihood in Equation 4. The only difference between the piecewise objective and the exact likelihood is in the second summation of Equation 7, which can be viewed as local approximation of the log partition function.

Finally, as an aside, the choice of pieces need not correspond to the graphical structure of the model. For example, in a linear chain MRF, which can be viewed as a weighted finite-state machine, we might choose to partition the state-transition diagram into pieces, and train each of these pieces separately. It is unclear if such training regimes are of practical interest, however.

Apart from its intuitive plausibility, another rationale for the piecewise estimator is provided by the following proposition:

**Proposition 1.** *For any set $\mathcal{P}$ of pieces, the piecewise approximation maximizes a lower bound on the likelihood, that is,*

$$A(\boldsymbol{\theta}) \leq \sum_{R \in \mathcal{P}} A_R(\boldsymbol{\theta}). \quad (8)$$

*Proof.* The bound is immediate upon expansion of $A(\boldsymbol{\theta})$.

$$A(\boldsymbol{\theta}) = \log \sum_{\mathbf{x}} \exp\left\{\sum_\alpha \theta_\alpha \phi_\alpha(\mathbf{x}_\alpha)\right\} \quad (9)$$

$$= \log \sum_{\mathbf{x}} \prod_{R \in \mathcal{P}} \exp\left\{\sum_{\alpha \in R} \theta_\alpha \phi_\alpha(\mathbf{x}_\alpha)\right\} \quad (10)$$

$$\leq \log \prod_{R \in \mathcal{P}} \sum_{\mathbf{x}_R} \exp\left\{\sum_{\alpha \in R} \theta_\alpha \phi_\alpha(\mathbf{x}_\alpha)\right\} \quad (11)$$

$$= \sum_{R \in \mathcal{P}} A_R(\boldsymbol{\theta}). \quad (12)$$

The bound from Equation 10 to Equation 11 is justified by considering the expansion of the product in equation Equation 11. The expansion contains every term of the summation in Equation 10, and all terms are nonnegative. □

### 3.1 APPLICATION TO CONDITIONAL RANDOM FIELDS

Piecewise estimation is especially well-suited for conditional random fields. As mentioned earlier, standard maximum-likelihood training for CRFs can require evaluating the instance-specific partition function $Z(\mathbf{x})$ for each training instance for each iteration of an optimization algorithm, which can be expensive even for linear chains. By using piecewise training, we need to compute only local normalization over small cliques, which for loopy graphs is potentially much more efficient.

If the training data is $\mathcal{D} = \{(\mathbf{x}^{(i)}, \mathbf{y}^{(i)})\}$, then the piecewise CRF objective function is

$$\ell_{\text{PW}}(\Lambda) = \sum_i \sum_{st} \sum_k \lambda_k f_k(y_s^{(i)}, y_t^{(i)}, \mathbf{x}^{(i)})$$
$$- \sum_i \sum_{st} A\left(\mathbf{x}^{(i)}; \Lambda\right), \quad (13)$$

where the local normalization factors are

$$A\left(\mathbf{x}^{(i)}; \Lambda\right) = \log \sum_{y_s, y_t} \exp\left\{\sum_k \lambda_k f_k(y_s, y_t, \mathbf{x}^{(i)})\right\}.$$

## 4 GENERALIZATIONS OF PIECEWISE TRAINING

In this section, we sketch another proof of Proposition 1, deriving it from the tree-reweighted bounds of Wainwright, Jaakkola, and Willsky [15], a connection which suggests generalizations of the simple piecewise training procedure. To simplify the exposition, in this section we assume that the pieces correspond to edges in a graphical model, but the ideas extend readily to more general pieces.

### 4.1 TREE-REWEIGHTED UPPER BOUNDS

Wainwright, Jaakkola, and Willsky [15] introduce a class of upper bounds on $A(\boldsymbol{\theta})$ that arise immediately from its convexity. The basic idea is to write the parameter vector $\boldsymbol{\theta}$ as a mixture of parameter vectors of tractable distributions, and then apply Jensen's inequality.

Let $\mathcal{T} = \{T_R\}$ be a set of tractable subgraphs of $\mathcal{G}$. For concreteness, think of $\mathcal{T}$ as the set of all spanning trees of $\mathcal{G}$; this is in fact the special case to which Wainwright, Jaakkola, and Willsky devote their attention. For each tractable graph $\mathcal{T}_R$, let $\boldsymbol{\theta}(T_R)$ be an exponential parameter vector that has the same dimensionality as $\boldsymbol{\theta}$, but *respects the structure* of $T_R$. More formally, this means that the entries of $\boldsymbol{\theta}(T_R)$ must be zero for edges that do not appear in $T_R$. Except for this, $\boldsymbol{\theta}(T_R)$ is arbitrary; there is no requirement that on its own, it matches $\boldsymbol{\theta}$ in any way.

Suppose we also have a distribution $\boldsymbol{\mu} = \{\mu_R | T_R \in \mathcal{T}\}$ over the tractable subgraphs, such that the original parameter vector $\boldsymbol{\theta}$ can be written as a combination of the per-tree parameter vectors:

$$\boldsymbol{\theta} = \sum_{T_R \in \mathcal{T}} \mu_R \boldsymbol{\theta}(T_R). \qquad (14)$$

In other words, we have written the original parameters $\boldsymbol{\theta}$ as a mixture of parameters on tractable subgraphs.

Then the upper bound on the log partition function $A(\boldsymbol{\theta})$ arises directly from Jensen's inequality:

$$A(\boldsymbol{\theta}) = A\left(\sum_{T_R \in \mathcal{T}} \mu_R \boldsymbol{\theta}(T_R)\right) \leq \sum_{T_R \in \mathcal{T}} \mu_R A(\boldsymbol{\theta}(T_R)). \qquad (15)$$

Because we have required that each graph $T$ be tractable, each term on the right-hand side of Equation 15 can be computed efficiently. If the size of $\mathcal{T}$ is large, however, then computing the sum is still intractable. We deal with this issue next.

A natural question about this bound is how to select $\boldsymbol{\theta}$ so as to get the tightest upper bound possible. For fixed $\boldsymbol{\mu}$, the optimization over $\boldsymbol{\theta}$ can be cast as a convex optimization problem:

$$\min_{\boldsymbol{\theta}} \sum_{T_R \in \mathcal{T}} \mu_R A(\boldsymbol{\theta}(T_R)) \qquad (16)$$

$$\text{s.t. } \boldsymbol{\theta} = \sum_{T_R \in \mathcal{T}} \mu_R \boldsymbol{\theta}(T_R). \qquad (17)$$

But this optimization problem can have astronomically many parameters, especially if $\mathcal{T}$ is the set of all spanning trees. The number of constraints, however, is much smaller, because the constraints are just one equality constraint for each element of $\boldsymbol{\theta}$. To collapse the dimensionality of the optimization problem, therefore, Wainwright, Jaakkola, and Willsky use the Lagrange dual of Equation 16, which can then be optimized using either standard optimization techniques, or a message passing algorithm similar to to BP. For our present purposes, however, it suffices to consider only the primal problem in Equation 16, which we use in the next section as a alternative derivation of piecewise bounds.

### 4.2 APPLICATION TO PIECEWISE UPPER BOUNDS

Now we discuss how the tree-reweighted upper bounds can be applied to piecewise training. As in the previous section, we will obtain an upper bound by writing the original parameters $\boldsymbol{\theta}$ as a mixture of tractable parameter vectors $\boldsymbol{\theta}(T)$. Consider the set $\mathcal{T}$ of tractable subgraphs induced by single edges of $\mathcal{G}$. Precisely, for each edge $E_R = (u_R, v_R)$ in $\mathcal{G}$, we add a (non-spanning) tree $T_R$ which contains all the original vertices but only the edge $E_R$. With each tree $T_R$ we associate an exponential parameter vector $\boldsymbol{\theta}(T_R)$.

Let $\boldsymbol{\mu}$ be a strictly positive probability distribution over edges. To use Jensen's inequality, we will need to have the constraint

$$\boldsymbol{\theta} = \sum_R \mu_R \boldsymbol{\theta}(T_R). \qquad (18)$$

Now, each parameter $\theta_i$ corresponds to exactly one edge of $\mathcal{G}$, which appears in only one of the $T_R$. Therefore, only one choice of subgraph parameter vectors $\{\boldsymbol{\theta}(T_R)\}$ meets the constraint (18), namely:

$$\boldsymbol{\theta}(T_R) = \frac{\boldsymbol{\theta}|_r}{\mu_R}, \qquad (19)$$

where $\boldsymbol{\theta}|_R$ is the restriction of $\boldsymbol{\theta}$ to $R$; that is, $\boldsymbol{\theta}|_R$ has the same entries and dimensionality as $\boldsymbol{\theta}$, but with zeros in all entries that are not included in the piece $R$.

Therefore, using Jensen's inequality, we immediately have the bound

$$A(\boldsymbol{\theta}) \leq \sum_R \mu_R A\left(\frac{\boldsymbol{\theta}|_R}{\mu_R}\right). \qquad (20)$$

This *reweighted piecewise* bound is clearly related to the basic piecewise bound in Equation 8, because $A(\boldsymbol{\theta}|_R)$ differs from $A_R(\boldsymbol{\theta})$ only by an additive constant which is independent of $\boldsymbol{\theta}$. In fact, a version of Proposition 1 can be derived by considering the limit of Equation 20 as $\boldsymbol{\mu}$ approaches a point mass on an arbitrary single piece $R^*$, but we do not present the details here.

The connection to the Wainwright et al. work suggests at least two generalizations of the basic piecewise method. The first is that the reweighted piecewise bound in Equation 20 can itself be minimized as an approximation to $A(\boldsymbol{\theta})$, yielding a variation of the basic piecewise method.

The second is that this line of analysis can naturally handle the case when pieces overlap. For example, in an MRF with both node and edge factors, we might choose each piece to be an edge factor with its corresponding node factors, hoping that this overlap will allow limited communication between the pieces which could improve the approximation. As long as there is a value of $\boldsymbol{\mu}$ for which the constraint in Equation 19 holds, then Equation 20 provides a bound we can minimize in an overlapping piecewise approximation.

In the experiments below, we evaluate weighting the pieces by a distribution $\mu$. We leave the exploration of overlapping pieces to future work.

## 5 EXPERIMENTS

The bound in Equation 8 is not tight. Because the bound does not necessarily touch the true likelihood at any point,

| Method | Overall F1 |
|---:|---|
| Piecewise | **91.2** |
| Pseudolikelihood | 84.7 |
| Per-edge PL | 89.7 |
| Exact | 89.9 |

Table 1: Comparison of piecewise training to exact and pseudolikehood training on a linear-chain CRF for named-entity recognition. On this tractable model, piecewise methods are more accurate than pseudolikelihood, and just as accurate as exact training.

| Method | Noun-phrase F1 |
|---:|---|
| Piecewise | **88.1** |
| Pseudolikelihood | 84.9 |
| Per-edge PL | 86.5 |
| BP | 86.0 |

Table 2: Comparison of piecewise training to other methods on a two-level factorial CRF for joint part-of-speech tagging and noun-phrase segmentation.

| Method | Token F1 | |
|---:|---|---|
| | location | speaker |
| Piecewise | **87.7** | 75.4 |
| Pseudolikelihood | 67.1 | 25.5 |
| Per-edge PL | 76.9 | 69.3 |
| BP | 86.6 | **78.2** |

Table 3: Comparison of piecewise training to other methods on a skip-chain CRF for seminar announcements.

maximizing it is not guaranteed to maximize the true likelihood. We turn to experiments to compare the accuracy of piecewise training both to exact estimation, and to other approximate estimators. A particularly interesting comparison is to pseudolikelihood, because it is a related local estimation method.

On three real-world natural language tasks, we compare piecewise training to exact ML training, approximate ML training using belief propagation, and pseudolikelihood training. To be as fair as possible, we compare to two variations of pseudolikelihood, one based on nodes and a structured version based on edges. Pseudolikelihood is normally defined as [1]:

$$\text{PL}(\boldsymbol{\theta}) = \prod_s p(x_s|\mathcal{N}(x_s)). \quad (21)$$

This objective function does not work well for sequence labeling, because it does not take into account strong interactions between neighboring sequence positions. In order to have a stronger baseline, we also compare to a per-edge version of pseudolikelihood:

$$\text{PL}_e(\boldsymbol{\theta}) = \prod_{st} p(x_s, x_t|\mathcal{N}(x_s, x_t)), \quad (22)$$

that is, instead of using the conditional distribution of each node, we use each edge, hoping to take more of the sequential interactions into account.

We evaluate piecewise training on three models used in previous work: a linear-chain CRF [8], a factorial CRF [13], and a skip-chain CRF [12]. All of these models use input features such as word identity, part-of-speech tags, capitalization, and membership in domain-specific lexicons; these are described fully in the original papers.

In all the experiments below, we optimize $\ell_{\text{PW}}$ using limited-memory BFGS. We use a Gaussian prior on weights to avoid overfitting. In previous work, the prior parameter had been tuned on each data set for belief propagation, and for the local models we used the same prior parameter without change. At test time, decoding is always performed using max-product belief propagation.

### 5.1 LINEAR-CHAIN CRF

First, we evaluate the accuracy of piecewise training on a tractable model, so that we can compare the accuracy to exact maximum-likelihood training. The task is named-entity recognition, that is, to find proper nouns in text. We use the CoNLL 2003 data set, consisting of 14,987 newswire sentences annotated with names of people, organizations, locations, and miscellaneous entities. We test on the standard development set of 3,466 sentences. Evaluation is done using precision and recall on the extracted chunks, and we report $F_1 = 2PR/P + R$. We use a linear-chain CRF, whose features are described elsewhere [10].

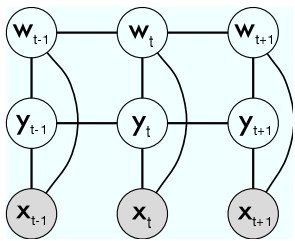

Figure 1: Graphical model for two-level FCRF for joint part-of-speech tagging and noun-phrase segmentation.

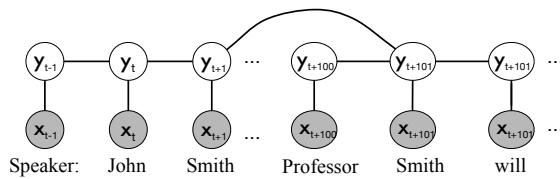

Figure 2: Graphical model for skip-chain CRF.

Piecewise training performs better than either of the pseudolikelihood methods. Even though it is a completely local training methods, piecewise training performs comparably to exact CRF training.

Now, in a linear-chain model, piecewise training has the same asymptotic complexity as exact CRF training, so we do not mean this experiment to advocate using the piecewise approximation for linear-chain graphs. Rather, that the piecewise approximation loses no accuracy on the linear-chain model is encouraging when we turn to loopy models, which we do next.

### 5.2 FACTORIAL CRF

The first loopy model we consider is the *factorial CRF* introduced by Sutton, Rohanimanesh, and McCallum [13]. Factorial CRFs are the conditionally-trained analogue of factorial HMMs [6]; it consists of a series of undirected linear chains with connections between cotemporal labels. This is a natural model for jointly performing multiple dependent sequence labeling tasks.

We consider here the task of jointly predicting part-of-speech tags and segmenting noun phrases in newswire text. Thus, the FCRF we use has a two-level grid structure, shown in Figure 1.

Our data comes from the CoNLL 2000 shared task [11], and consists of sentences from the Wall Street Journal annotated by the Penn Treebank project [9]. We consider each sentence to be a training instance, with single words as tokens. We report results here on subsets of 223 training sentences, and the standard test set of 2012 sentences. Results are averaged over 5 different random subsets. There are 45 different POS labels, and the three NP labels. We report F1 on noun-phrase chunks.

In previous work, this model was optimized by approximating the partition function using belief optimization, but this was quite expensive. Training on the full data set of 8936 sentences required about 12 days of CPU time.

Results on this loopy data set are presented in Table 2. Again, the piecewise estimator performs better than either version of pseudolikelihood and maximum-likelihood esti-mation using belief propagation.

### 5.3 SKIP-CHAIN CRF

Finally, we consider a model with many irregular loops, which is the skip chain model introduced by Sutton and McCallum [12]. This model incorporates certain long-distance dependencies between word labels into a linear-chain model for information extraction.

The task is to extract information about seminars from email announcements. Our data set is a collection of 485 e-mail messages announcing seminars at Carnegie Mellon University. The messages are annotated with the seminar's starting time, ending time, location, and speaker. This data set is due to Dayne Freitag [5], and has been used in much previous work.

Often the speaker is listed multiple times in the same message. For example, the speaker's name might be included both near the beginning and later on, in a sentence like "If you would like to meet with Professor Smith..." It can be useful to find both such mentions, because different information can be in the surrounding context of each mention: for example, the first mention might be near an institution affiliation, while the second mentions that Smith is a professor.

To increase recall of person names, we wish to exploit that when the same word appears multiple times in the same message, it tends to have the same label. In a CRF, we can represent this by adding edges between output nodes $(y_i, y_j)$ when the words $x_i$ and $x_j$ are identical and capitalized. An example of this is shown in Figure 2. Thus, the conditional distribution $p(\mathbf{y}|\mathbf{x})$ has different graphical structure for different input configurations $\mathbf{x}$.

Consistently with the previous work on this data set, we use 10-fold cross validation with a 50/50 training/test split. We report per-token F1 on the speaker and location fields, the most difficult of the four fields. Most documents contain many crossing skip-edges, so that exact maximum-likelihood training using junction tree is completely infeasible, so instead we compare to approximate training using loopy belief propagation.

Results on this model are given in Table 3. Pseudolikelihood performs particularly poorly on this model. Piecewise

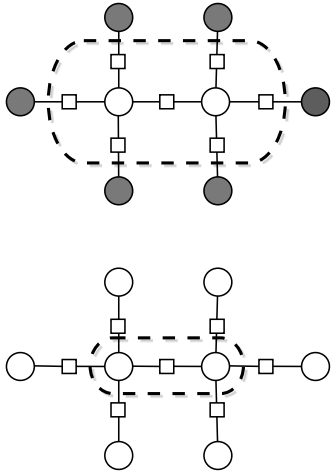

Figure 3: Schematic factor-graph depiction of the difference between pseudolikelihood (top) and piecewise training (bottom). Each term in pseudolikelihood normalizes the product of many factors (as circled), while piecewise training normalizes over one factor at a time.

| Model | Basic | Reweighted |
|---:|---|---|
| Linear-chain | 91.2 | 90.4 |
| FCRF | 88.1 | 86.4 |
| Skip-chain (location) | 87.7 | 75.5 |
| Skip-chain (speaker) | 75.4 | 69.2 |

Table 4: Comparison of basic piecewise training to reweighted piecewise bound with uniform $\mu$.

estimation performs much better, but worse than approximate training using BP.

Piecewise training is faster than loopy BP: in our implementation piecewise training used on average 3.5 hr, while loopy BP used 6.8 hr. To get these loopy BP results, however, we must carefully initialize the training procedure: We initialize the linear-chain part of the skip-chain from the weights of a fully-trained linear-chain CRF. If we instead start at the uniform distribution, not only does loopy BP training take much longer, over 10 hours, but testing performance is much worse, because the convex optimization procedure has difficulty with noisier gradients. With uniform initialization, loopy BP does not converge for all training instances, especially at early iterations of training. Carefully initializing the model parameters seems to alleviate these issues, but this model-specific tweaking was unnecessary for piecewise training.

,

### 5.4 REWEIGHTED PIECEWISE TRAINING

We also evaluate a reweighted piecewise training, a modification to the basic piecewise estimator discussed in Section 4, in which the pieces are weighted by a convex combination (Equation 20). The performance of reweighted piecewise training with uniform $\mu_R$ is presented in Table 4. In all cases, the reweighted piecewise method performs worse than the basic piecewise method. What seems be happening is that in each of these models, there are several hundred edges, so that the weight $\mu_R$ for each region is rather small, perhaps around 0.01. For each piece $R$, reweighted bound includes a term $A\left(\boldsymbol{\theta}|_R/\mu_R\right)$. If $\mu_R$ is around 0.01, then this means that we multiply the log factor values by 100 before evaluating $A$. This multiplier is so extreme that the term $A\left(\boldsymbol{\theta}|_R/\mu_R\right)$ is dominated by maximum-value weight in $\boldsymbol{\theta}|_R$.

## 6 RELATED WORK

Because the piecewise estimator is such an intuitively appealing method, it has been used in several scattered places in the literature, for tasks such as information extraction [17], collective classification [7], and computer vision [4]. In these papers, the piecewise method is reported as a successful heuristic for training large models, but its performance is not compared against other training methods. We are unaware of previous work systematically studying this procedure in its own right.

As mentioned earlier, the most closely related procedure that has been studied statistically is pseudolikelihood [1, 2]. The main difference is that piecewise training does not condition on neighboring nodes, but ignores them altogether during training. This is depicted schematically by the factor graphs in Figure 3. In pseudolikelihood, each locally-normalized term for a variable or edge in pseudolikelihood includes contributions from a number of factors that connect to the neighbors whose observed values are taken from labeled training data. All these factors are circled in the top section of Figure 3. In piecewise training, each factor becomes an independently, locally-normalized term in the objective function.

## 7 CONCLUSION

In this paper, we study piecewise training, an intuitively appealing procedure that separately trains disjoint pieces of a loopy graph. We show that this procedure can be justified as maximizing a loose bound on the log likelihood. On three real-world language tasks with different model structures, piecewise training outperforms several versions of pseudolikelihood, a traditional local training method. On two of the data sets, in fact, piecewise training is more accurate than global training using belief propagation.

Many properties of piecewise training remain to be explored. Our results indicate that in some situations piecewise training should replace pseudolikelihood as the local training method of choice. Characterizing the situations in which piecewise is preferable to pseudolikelihood, and vice versa, is an important avenue of future work. In particular, the experiments here all used conditional training, which make local training easier because of the large amount of information in the conditioning variables. In generative training, there may be much less local information, making piecewise training much less effective.

## Acknowledgements

We thank Louis Theran and Tommi Jaakkola for useful discussions, and the anonymous reviewers for very helpful comments. This work was supported in part by the Center for Intelligent Information Retrieval, in part by the National Science Foundation under NSF grants #IIS-0326249 and #IIS-0427594. Any opinions, findings and conclusions or recommendations expressed in this material are the author(s) and do not necessarily reflect those of the sponsor.